\documentclass[a4paper]{article}
\usepackage{INTERSPEECH2021}
\usepackage{float}
\usepackage{hyperref}
\pdfoutput=1

\title{Speak or Chat with Me: \\End-to-End Spoken Language Understanding System with Flexible Inputs}
\name{Sujeong Cha$^{1*}$, Wangrui Hou$^{1*}$, Hyun Jung$^{1*}$, My Phung$^{1*}$\thanks{* The authors contributed equally.}, Michael Picheny$^1$, \\Hong-Kwang Kuo$^2$, Samuel Thomas$^2$, Edmilson Morais$^3$}

\address{
  $^1$New York University, USA\\
  $^2$IBM Research AI, USA   $^3$IBM Research AI, Brazil}
\email{\{sjc433, wh916, hj1399, mtp363, map22\}@nyu.edu,\\ \{hkuo, sthomas\}@us.ibm.com, edmorais@br.ibm.com}

\begin{document}

\maketitle

\begin{abstract}
A major focus of recent research in spoken language understanding (SLU) has been on the end-to-end approach where a single model can predict intents directly from speech inputs without intermediate transcripts. However, this approach presents some challenges. First, since speech can be considered as personally identifiable information, in some cases only automatic speech recognition (ASR) transcripts are accessible. Second, intent-labeled speech data is scarce. To address the first challenge, we propose a novel system that can predict intents from flexible types of inputs: speech, ASR transcripts, or both. We demonstrate strong performance for either modality separately, and when both speech and ASR transcripts are available, through system combination, we achieve better results than using a single input modality. To address the second challenge, we leverage a semantically robust pre-trained BERT model and adopt a cross-modal system that co-trains text embeddings and acoustic embeddings in a shared latent space. We further enhance this system by utilizing an acoustic module pre-trained on LibriSpeech and domain-adapting the text module on our target datasets. Our experiments show significant advantages for these pre-training and fine-tuning strategies, resulting in a system that achieves competitive intent-classification performance on Snips SLU and Fluent Speech Commands datasets.

\end{abstract}

\section{Introduction}
Traditional spoken language understanding (SLU) systems use a cascaded automatic speech recognition (ASR) system to convert input speech signals into transcripts, followed by a natural language understanding (NLU) model to predict the intent from text transcripts \cite{Lee10-RAT}-\cite{Haghani18-FAT}. While this approach has been widely adopted, there are significant limitations. The intermediate step of mapping audio to text is expensive to train and creates errors in the transcripts. Such errors greatly affect the final intent classification. Additionally, since these two models are trained independently, the primary metric of interest (intent classification accuracy) cannot be directly optimized. Due to this problem, end-to-end (E2E) SLU models that directly map a speech signal input to an SLU output have become popular \cite{Radfar20-ETE}-\cite{Lugosch19-SMP}.

However, this approach also presents some challenges. First, in some cases, only automatic speech recognition (ASR) transcripts are available since speech is considered as personally identifiable information and may not be stored. This implies that deployable E2E systems must still be able to handle text (ASR) input. Second, an E2E approach requires a large amount of data to produce comparable results to traditional cascaded ASR-NLU systems. This can be addressed by using pre-training to reduce the amount of training data required. For example, researchers have pre-trained models on large ASR datasets such as LibriSpeech \cite{Lugosch19-SMP}\cite{Panayotov15-LAA} to relax audio data requirements, and have used pre-trained BERT networks \cite{Rongali20-ETL}-\cite{Devlin18-BPT} to relax text data requirements.

One way to address both challenges is described in \cite{Rongali20-ETL}. A single architecture is used to jointly train an E2E system across a variety of tasks including ASR, SLU, masked LM prediction, and hypothesis prediction from text. Alternatively, authors in \cite{Huang20-LUT} successfully co-train a text-to-intent (T2I) model and speech-to-intent (S2I) model to closely align the acoustic embeddings with BERT-based text embeddings. Further improving on this cross-modal approach, authors in \cite{Agrawal20-TYE} employ the triplet loss function to learn a robust cross-modal latent space. Training the acoustic and text embeddings in the shared latent space makes it easier to combine separate audio and text data.  

In this paper, we propose an architecture that extends the architectures in \cite{Huang20-LUT}\cite{Agrawal20-TYE} because we believe that the common latent space approach with \hyperref[sec:triplet-loss]{triplet loss} is an easy but powerful way to utilize discriminative approaches in the training process. We also broaden the application scenarios of our system to perform intent classification with ASR transcripts to address the first challenge - when original audio files are not available due to privacy reasons. This also allows us to produce an intent prediction for a chat transcript without the audio counterpart. We choose the cross-modal system in \cite{Agrawal20-TYE} as our baseline model and further investigate and improve upon their approach. Their model consists of two branches: an acoustic branch that takes audio input, and a BERT text branch that takes the ground truth transcripts of the audio files. Triplet loss ties the acoustics embeddings and text embeddings together, thus forcing samples in the same class to be closer in the feature space and pushing samples from different classes to be further apart \cite{Agrawal20-TYE}.  

\begin{figure*}[htb!]
  \centering
  \includegraphics[scale=0.35]{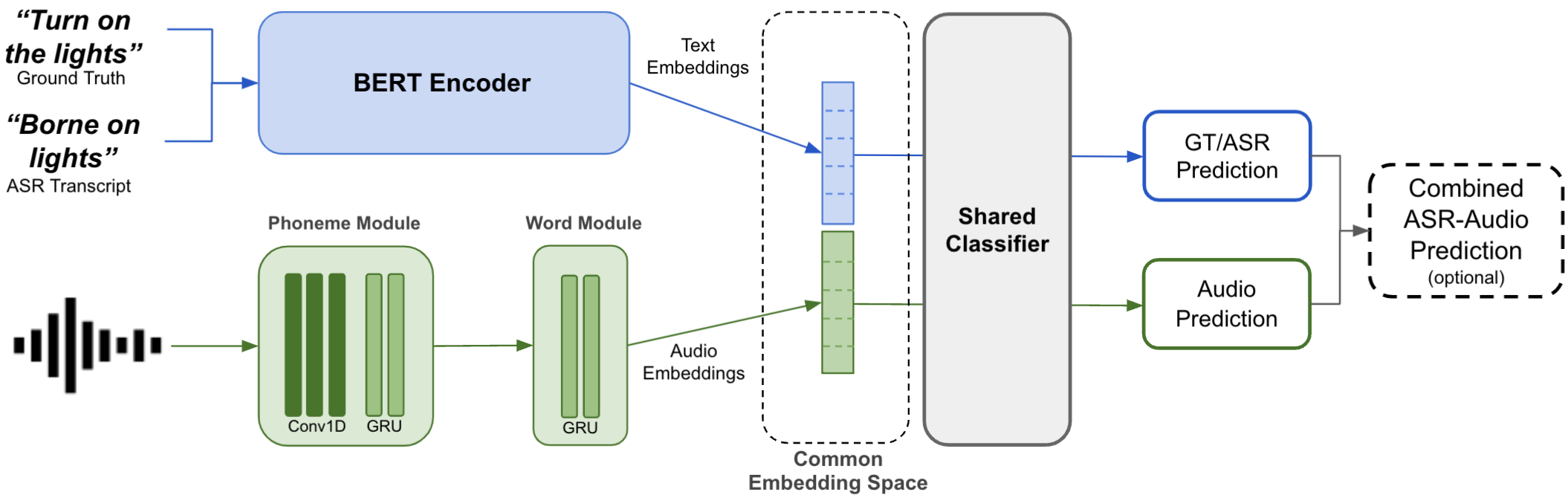}
  \caption{Diagram of ASR-Text-Speech Model.}
  \label{fig:e2e_ast}
\end{figure*}

The novelties we introduce to the baseline model include 1) improvements to the acoustic and text branches by pre-training the acoustic branch and domain-adapting the BERT text branch, 2) the ability to take flexible types of input data (i.e. speech, ASR-transcripts/chat, or both), and 3) system combination \cite{Kittler98-OCC}\cite{Price20-IET} to combine intent predictions from acoustic and text branches, which performs better than an individual branch. We obtain significant improvements in performance over the baseline model on the Snips \cite{Saade18-SLU} and  Fluent Speech Commands (FSC) datasets \cite{Lugosch19-SMP}. To welcome researchers to improve upon our work similar to \cite{Lugosch19-SMP}\cite{Agrawal20-TYE}, we are releasing our codebase.\footnote{https://github.com/CoraJung/flexible-input-slu} 

\section{Model}
A major challenge when training an E2E SLU model is the scarcity of publicly available SLU datasets with paired utterance audio and semantic labels. ASR and NLU specific training data are much more accessible. Our approaches to overcome this challenge are 1) creating a tied cross-modal latent space that enables us to co-train the acoustic and text embeddings and 2) leveraging ASR and NLU data to pre-train the acoustic and text modalities independently before fine-tuning the joint model. Our cross-modal SLU system consists of two branches: an acoustic branch that takes audio input to produce acoustic embeddings and a text branch that takes in either ground truth or ASR transcripts of the audio files to produce text embeddings, as shown in Figure~\ref{fig:e2e_ast}.

The acoustic branch is pre-trained on LibriSpeech, and the text branch is a pre-trained BERT model. The two embeddings are tied together in a common embedding space through a triplet loss. The two branches share the final classification layer that is trained to predict intent classes from acoustic embeddings, text embeddings, or both. Unlike the baseline model, our goal is to achieve higher audio accuracy while preserving the performance of the pre-trained text branch. The final input to the model training can include either 1) ground truth text transcripts and speech (Text-Speech Model) or 2) ASR transcripts, ground truth transcripts, and speech (ASR-Text-Speech Model). During inference time, the model can flexibly perform intent classification from any of these three types of input: 1) speech, 2) ASR/ground truth transcripts, and 3) speech combined with ASR/ground truth transcripts. In practice, an example of ground truth transcripts can be a chat utterance typed by a user.

\subsection{BERT Text Branch}

We follow the text branch architecture of the baseline model, which has a pre-trained BERT-base-cased model \cite{Wolf19-Hts} as the text encoder. The text branch takes either ground truth or ASR-generated transcripts to create text embeddings. To boost the performance of our text branch, we also experiment with domain-adapting the text branch on our target datasets before co-training it with the acoustic branch.

\subsection{LibriSpeech Pre-trained Audio Branch}

The acoustic branch of the baseline model is a multi-layer Bi-LSTM network with random initialization that computes acoustic embeddings based on audio input. Because this acoustic branch functions similarly to an ASR model, it can greatly benefit from pre-training on an ASR-specific dataset, especially when E2E SLU training data is generally scarce \cite{Lugosch19-SMP}. Therefore, we directly utilize the LibriSpeech pre-trained phoneme and word modules from \cite{Lugosch19-SMP} for our acoustic branch. The phoneme module processes raw input waveform and outputs a hidden representation of the phonemes, and the word module takes the phoneme embeddings as input and produces word embeddings. Ultimately, the word embeddings output from the word module is passed through a linear layer to project the final acoustic embeddings into the same embedding space as the text embeddings. 

\subsection{Shared Embedding Space and Triplet Loss}\label{sec:triplet-loss}

The key component of the cross-modal architecture is the shared embedding space that allows the acoustic embeddings to "learn" from the text embeddings. This objective is achieved through a triplet loss function, denoted $L_e$:
\begin{equation}
  L_e(x_a,x_{+},x_{-}) = max(0,m + d(x_a,x_{+}) - d(x_a,x_{-}))
  \label{eq1}
\end{equation}
where $x_a$ represents an acoustic embedding of the current sample, $x_{+}, x_{-}$ represents text embeddings of the positive and negative sample, $m$ represents the margin, and $d$ represents the squared distance. The benefit of this triplet loss is to move the acoustic embedding of the current example closer to the text embedding of the positive example and away from that of the negative example \cite{Agrawal20-TYE}. In brief, the triplet loss uses three inputs: one acoustic embedding of the current training example and two text embeddings coming from one negative and one positive example. A positive example is drawn randomly from the same intent class as the current example and a negative example is chosen arbitrarily from other intent classes in the training data. Different from the baseline model, our acoustic embedding of the current example will be the word embedding output from the word module in the acoustic branch. The text embeddings of positive and negative examples are outputs of the BERT encoder.

\subsection{Model Selection Process}

The final loss is a weighted sum of the triplet loss and cross-entropy losses from audio- and text-based predictions of the intents obtained from the last classification layer, which are denoted $L_{cls}^{acoustic}$ and $L_{cls}^{text}$ respectively in the equation below:
\begin{equation}
  L = L_{cls}^{acoustic} + \lambda_1L_{cls}^{text} + \lambda_2L_e
  \label{eq2}
\end{equation}
where $\lambda_1, \lambda_2$ are weights associated with the loss components.
Originally, the baseline model chooses the best model based on the acoustic branch's validation accuracy only, which causes the text prediction accuracy to degrade. To preserve the performance of the text branch, we select our best model based on the simple average of validation accuracy from both acoustic and text branches. This enables us to improve the performance of the text branch compared to the baseline model and ultimately allows the model to perform well on not only speech but also ASR-transcript inputs.

\section{Experiments}

\subsection{Datasets}
All of our experiments are performed on two publicly available datasets, namely Snips and FSC.\\

\noindent\textbf{Snips SLU}  We use the “SmartLights" close-field subset of the Snips SLU dataset, which contains English spoken commands related to smart light appliances, such as “Can you turn on the light in the kitchen, please?”, which are paired with intent labels, such as “TurnOnLight” \cite{Saade18-SLU}. It contains 1,660 unique utterances spoken by 21 speakers, mapping to six unique intent labels. To make our results comparable, we follow the data preparation procedures used in \cite{Agrawal20-TYE}. In particular, the dataset is split into training (80\%), validation (10\%), and test (10\%) sets.\\

\noindent\textbf{Fluent Speech Commands}  The FSC dataset contains 30,043 spoken English commands for a smart home or virtual assistant, such as “turn on the kitchen lights”. It comes with full-text transcripts and intent labels in the form of “action”, “object”, “location”. There are 248 unique transcripts mapping to 31 unique intents. We remove a total of 28 utterances in the training, validation, and test sets that have empty audio and obtain a final dataset of 30,015 utterances \cite{Lugosch19-SMP}.\\

\noindent\textbf{ASR Transcript Generation}  The Snips and FSC datasets only contain ground truth transcripts; no ASR transcripts are provided. We generate ASR transcripts by passing the audio inputs to an existing ASR model trained on LibriSpeech using the Kaldi Speech Recognition toolkit \cite{Povey11-TKS}. The average word-error-rate (WER) for generating ASR transcripts for Snips and FSC data are 34.1\% and 25.6\% respectively. Note that we use an off-the-shelf LibriSpeech system without any adaptation because we want to specifically examine the robustness of our system in the presence of relatively noisy ASR transcripts.

\subsection{Training Setup}
We first present the training setup for the Text-Speech model, trained on ground truth transcripts and speech. This model serves as the groundwork upon which we build our best models, ASR-Text-Speech-1 (without text encoder domain adaptation) and ASR-Text-Speech-2 (with such adaptation). Note that we perform hyper-parameter tuning for learning rates and set all other hyper-parameters the same as \cite{Lugosch19-SMP} and \cite{Agrawal20-TYE} to make the results comparable. \\ 

\noindent\textbf{Text-Speech Model}  As explained in Section 2.1, a pre-trained BERT-base-cased model \cite{Wolf19-Hts} with default configurations (12 layers, 768 final embedding dimension) acts as our text encoder. The acoustic branch has phoneme and word modules (three Conv1D layers; four GRU layers, 128 hidden units/layer) pre-trained on LibriSpeech, and one final linear layer (768 final embedding dimension). The two branches share the same final linear classifier and the output dimension varies according to the number of unique labels in training data. All pre-trained modules (BERT encoder, phoneme module, and word module) are then fine-tuned on our target dataset. The learning rate for both the acoustic encoder and the final classifier is 1e-3, and the learning rate for the BERT encoder is 2e-5.\\

\noindent\textbf{ASR-Text-Speech-1 Model}  ASR-Text-Speech-1 builds upon the Text-Speech model. One major change is that the training data for the text encoder of ASR-Text-Speech-1 includes both ground truth and ASR transcripts, while Text-Speech is only trained on ground truth transcripts. Furthermore, our experiments have shown that, for ASR-Text-Speech-1, freezing the phoneme module of the acoustic branch produces better performance than fine-tuning all modules.\\

\noindent\textbf{ASR-Text-Speech-2 Model}
\label{sec:ATS2} Similar to the ASR-Text-Speech-1 model, ASR-Text-Speech-2 also improves upon Text-Speech by taking ASR transcripts as its input. However, ASR-Text-Speech-2 uses ASR transcripts in addition to ground truth transcripts to domain-adapt the BERT branch before joint training it with the acoustic branch. In this domain adaptation phase, we build a separate BERT model that has a structure identical to the text encoder in Text-Speech. We domain-adapt this BERT model on the ground truth and ASR transcripts from our target datasets with a learning rate of 2e-5. We then use this domain-adapted BERT model as our text branch without any further fine-tuning. By freezing the BERT text branch in the final training round, we let the domain-adapted text embeddings guide our audio embeddings. This is why, in this stage, we only use pairs of audio and ground truth transcripts, which provides better-quality text embeddings compared to ASR transcripts.\\

\noindent\textbf{System Combination}  During inference, when both speech and ASR transcripts are available, we use system combination to leverage the ``opinions" of the two branches, as shown in Figure~\ref{fig:e2e_ast}, and achieve a better performance than using the prediction from each branch. This combined prediction is an element-wise average of the classifier's outputs (probabilistic predictions for each class label) from the acoustic branch and the text branch. Moreover, considering the flexible nature of our inputs, system combination helps our SLU system to consistently produce one single prediction. In summary, this system combination process ensures that 1) we achieve more accurate predictions and 2) our system can produce a single output when receiving both speech and ASR transcripts.

\section{Results}

\subsection{Snips Results}

We first apply our models to the Snips dataset. The \textbf{GT}, \textbf{Audio}, and \textbf{ASR} columns of Table 1 show the intent prediction accuracy when the models are tested on ground truth transcripts, audio input, and ASR transcripts, respectively. The \textbf{Combined} column shows the intent accuracy from system combination which combines the predictions of the acoustic and text branch to produce a final intent prediction. We also take the simple average of audio, ASR, and combined accuracy for each model to create the \textbf{Average} column. This column serves as an indicator of a model’s overall performance, making comparisons easier and cleaner.

\begin{table}[th]
  \caption{Snips Results, Accuracy (\%)}
  \label{tab:Snips}
  \centering
\renewcommand{\arraystretch}{1.2}
\resizebox{\columnwidth}{!}{
\begin{tabular}{cccccc}
\toprule
\textbf{Model}    & \textbf{GT} & \textbf{Audio} & \textbf{ASR} & \textbf{Combined} & \textbf{Average} \\
\midrule
\textbf{Agrawal et.al. \cite{Agrawal20-TYE}} & -    &69.88& -    & -    &- \\
\textbf{Rongali et.al. \cite{Rongali20-ETL}} & -    &84.88\textsuperscript{*} & -    & -    &- \\
\textbf{Text-Speech}         &98.19 &75.90 &78.31 &87.95 &80.72 \\
\textbf{ASR-Text-Speech-1}   &97.59 &78.31 &83.73 &89.76 &83.93\\
\textbf{ASR-Text-Speech-2}   &97.59 &80.12 &80.72 &86.75 &82.53 \\
\bottomrule
\end{tabular}}
\footnotetext\footnotesize\scriptsize{*Hard to compare because run on different (unpublished) train and test sets.}
\end{table}

Based on the results, our three proposed models, which replace the unpre-trained LSTM-based acoustic branch in [14] with the GRU-based acoustic branch pre-trained on LibriSpeech in [10], outperform the baseline model on audio accuracy. Since the baseline model only generates predictions based on audio input, the audio accuracy is the only category where we can compare our results with the baseline. To be comprehensive, we also included results from \cite{Rongali20-ETL}. However, \cite{Rongali20-ETL} utilizes a larger training set (including Snips far-field subset) and a different test set from ours, which makes direct comparisons difficult.

Among our three models, ASR-Text-Speech-1 has the highest average accuracy on Snips. Moreover, by comparing our two ASR-Text-Speech models with the Text-Speech model, we observe that adding ASR transcripts to our training data improves our model's performances on audio, ASR, and combined accuracy. There are only trivial differences in the results of our two ASR-Text-Speech models, which implies that applying domain adaptation to our text branch does not provide notable benefits to our model. Compared to ASR-Text-Speech-1, ASR-Text-Speech-2 performs better on audio input but worse on ASR transcripts. This result confirms our hypothesis in \hyperref[sec:ATS2]{3.2} that fixing the text branch after domain-adaptation guides audio embeddings better; however, this comes with a trade-off in text predictions because the text branch is not fine-tuned in the final training stage. In summary, we would recommend the ASR-Text-Speech-1 based on its competitive performance and simplicity in training setup. 

As expected, system combination (the \textbf{Combined} column) generally achieves higher accuracy than using acoustic or ASR transcript information alone for prediction. 

\subsection{FSC Results}

\begin{table}[th]
  \caption{FSC Results, Accuracy (\%)}
  \label{tab:FSC}
  \centering
\renewcommand{\arraystretch}{1.2}
\resizebox{\columnwidth}{!}{
\begin{tabular}{cccccc}
\toprule
\textbf{Model}    & \textbf{GT} & \textbf{Audio} & \textbf{ASR} & \textbf{Combined} & \textbf{Average} \\
\midrule
\textbf{Agrawal et.al. \cite{Agrawal20-TYE}} & -     &95.65 & -    & -    &- \\
\textbf{Lugosch et.al. \cite{Lugosch19-SMP}} & -     &98.75 & -    & -    &-  \\
\textbf{Rongali et.al. \cite{Rongali20-ETL}} & -     &99.50 & -    & -    &-  \\
\textbf{Text-Speech}         &100.00 &99.13 &87.92 &97.36 &94.80 \\
\textbf{ASR-Text-Speech-1}   &100.00 &99.18 &98.18 &99.45 &98.94\\
\textbf{ASR-Text-Speech-2}   &100.00 &98.95 &98.23 &99.53 &98.90 \\
\bottomrule
\end{tabular}}
\end{table}

\noindent Next, we test the efficacy of our proposed models on the FSC dataset (Table 2). Similar to the results on Snips, our models perform well in all four categories and ASR-Text-Speech-1 is still our best performing model. Additionally, our models give competitive results, outperforming results reported in \cite{Lugosch19-SMP} and \cite{Agrawal20-TYE}. Our models perform slightly worse than the model proposed in \cite{Rongali20-ETL} since their model is pre-trained with additional SLU datasets such as their internal datasets.

Also notice that we observe a larger improvement in ASR accuracy from Text-Speech model to ASR-Text-Speech models in FSC (around 10\%) than in Snips (around 2-5\%). We hypothesize that, because ASR transcripts of FSC has a lower WER than those of Snips, the addition of better-quality ASR transcripts benefits FSC more during training.

\subsection{Additional Results}

\begin{table}[th]
  \caption{Snips Text Branch Results, Accuracy (\%)}
  \label{tab:Text-only}
  \centering
\renewcommand{\arraystretch}{1.2}
\scalebox{0.8}{
\begin{tabular}{ccc}
\toprule
\textbf{Model}    & \textbf{GT} &  \textbf{ASR} \\
\midrule
\textbf{Text-Only}     & 95.18 & 77.11 \\
\textbf{ASR-Text}    & 95.78 & 78.92 \\
\textbf{ASR-Text-Speech-1} & 97.59 & 83.73  \\
\bottomrule
\end{tabular}}
\end{table}

\noindent We also examine the impact of varying the input type on the performance of our text branch using Snips (Table 3) and FSC. The \textbf{Text-Only} model is a pre-trained BERT-base-cased model fine-tuned on only the ground truth transcripts, and the \textbf{ASR-Text} model is the same BERT model fine-tuned on both the ground truth and the ASR transcripts. By comparing the Text-Only model with the ASR-Text model, we further provide evidence that adding ASR transcripts in the training stage improves model performance on both ground truth and ASR test sets. Interestingly, adding the acoustic branch to the ASR-Text model further improves the performance of our text branch when tested on either ground truth or ASR transcripts. Similar results are observed when we run these experiments on FSC.

\section{Conclusion}
In general, we can clearly see the improvements achieved by the novelties proposed in this paper.

First, our ASR-Text-Speech models can predict intents from flexible types of inputs: speech, ASR transcripts (or chat utterances), or a combination of both. Our model gives excellent performance when only the ASR transcripts are available to predict intents. If both speech and ASR transcripts are available at test time, combining the intent probability predictions of the acoustic and text branches gives more accurate final predictions. Between our two versions of our proposed model, ASR-Text-Speech-1 has better overall predictions and is also more compact and efficient compared to ASR-Text-Speech-2 due to the absence of domain-adapting BERT on target datasets in the pre-training round. 

Second, pre-training on LibriSpeech gives our system a more powerful acoustic branch, enabling our model to outperform the baseline. Furthermore, by saving the best model based on both audio- and text-based prediction accuracy, our model preserves the performance of the text branch compared to the baseline model. It is important to note that co-training the text branch and the acoustic branch does not only improve the acoustic branch’s performance, which is the original motivation, but also boosts the performance of the text branch. We hypothesize that since the shared classifier between the two modalities is trained to perform predictions on either text or acoustic embeddings, it may be better regularized and more robust to unseen or noisy data at test time. This also explains why adding noisy ASR transcripts improves the results of our Text-Only model.

\section{Acknowledgments}

We would like to thank Markus Mueller from Amazon Alexa Machine Learning and Loren Lugosch from McGill University and Mila AI Research Institute for releasing their codebases and answering our questions. We would also like to thank Amber Wang from New York University for generating the ASR transcripts for Snips SLU and FSC.

\bibliographystyle{IEEEtran}

\end{document}